\documentclass[11pt,a4paper]{article}
\usepackage[hyperref]{emnlp-ijcnlp-2019}
\usepackage{times}
\usepackage{latexsym}
\usepackage{url}
\usepackage{array}
\usepackage{color}
\usepackage{bm}
\usepackage{enumitem}
\usepackage{booktabs}
\usepackage{amsmath}
\usepackage{amssymb}
\usepackage{amsfonts}
\usepackage{multirow}
\usepackage{subcaption}
\usepackage{graphicx}
\usepackage{bbding}
\usepackage{pifont}
\usepackage{wasysym}
\usepackage{microtype}  %
\usepackage[linesnumbered,ruled,vlined]{algorithm2e}

\newcolumntype{P}[1]{>{\centering\arraybackslash}p{#1}}

\newcommand{\mask}{$ \langle \textit{mask} \rangle $}

\newcommand{\betanoise}{\ensuremath{\beta_{\mathrm{random}}}}

\newcommand{\backtrans}{\textsc{Backtrans (noisy)}}
\newcommand{\samplebacktrans}{\textsc{Backtrans (sample)}}
\newcommand{\directnoise}{\textsc{DirectNoise}}
\newcommand{\joint}{\textsc{Joint}}
\newcommand{\pretrain}{\textsc{Pretrain}}
\newcommand{\spell}{\textsc{SSE}}
\newcommand{\pseudodata}{\ensuremath{\mathcal{D}_{p}}}
\newcommand{\genuinedata}{\ensuremath{\mathcal{D}_{g}}}
\newcommand{\seedcorpus}{\ensuremath{\mathcal{T}}}
\newcommand{\fscore}{\ensuremath{\mathrm{F}_{0.5}}}

\newcommand{\todo}[1]{}
\renewcommand{\todo}[1]{{\color{red} TODO: {#1}}}

\aclfinalcopy %

\title{An Empirical Study of Incorporating Pseudo Data\\into Grammatical Error Correction}

\author{
  Shun Kiyono$^{\,1,2}$ ~ Jun Suzuki$^{\,2,1}$ ~ Masato Mita$^{\,1,2}$ ~ Tomoya Mizumoto$^{\,1,2}$\thanks{\; Current affiliation: Future Corporation}  ~ Kentaro Inui$^{\,2,1}$ \\
    ${}^{1}$ RIKEN Center for Advanced Intelligence Project ~~ ${}^{2}$ Tohoku University \\
   \texttt{\{shun.kiyono, masato.mita, tomoya.mizumoto\}@riken.jp;} \\ 
   \texttt{\{jun.suzuki,inui\}@ecei.tohoku.ac.jp} \\ 
}

\date{}

\begin{document}
\maketitle
\begin{abstract}
The incorporation of pseudo data in the training of grammatical error correction models has been one of the main factors in improving the performance of such models.
However, consensus is lacking on experimental configurations, namely, choosing how the pseudo data should be generated or used. 
In this study, these choices are investigated through extensive experiments, and state-of-the-art performance is achieved on the CoNLL-2014 test set ($\fscore{}=65.0$) and the official test set of the BEA-2019 shared task ($\fscore{}=70.2$) without making any modifications to the model architecture.
\end{abstract}

\section{Introduction}
To date, many studies have tackled grammatical error correction (GEC) as a machine translation (MT) task, in which ungrammatical sentences are regarded as the source language and grammatical sentences are regarded as the target language.
This approach allows cutting-edge neural MT models to be adopted.
For example, the encoder-decoder (EncDec) model~\citep{sutskever:2014:NIPS,bahdanau:2015:ICLR}, which was originally proposed for MT, has been applied widely to GEC and has achieved remarkable results in the GEC research field~\citep{ji:2017:nested,chollampatt:2018:AAAI,junczys:2018:NAACL}.

However, a challenge in applying EncDec to GEC is that EncDec requires a large amount of training data~\citep{koehn:2017:NMT}, but the largest set of publicly available parallel data (Lang-8) in GEC has only two million sentence pairs~\citep{mizumoto:2011:IJCNLP}.
Consequently, the method of augmenting the data by incorporating pseudo training data has been studied intensively~\citep{xie:2018:NAACL,ge:2018:ACL,lichtarge2019corpora,zhao2019improving}.

When incorporating pseudo data, several decisions must be made about the experimental configurations, namely, (i) the method of generating the pseudo data, (ii) the seed corpus for the pseudo data, and (iii) the optimization setting (Section~\ref{sec:problem-formulation}).
However, consensus on these decisions in the GEC research field is yet to be formulated.
For example, \citet{xie:2018:NAACL} found that a variant of the backtranslation~\citep{sennrich:2016:backtrans} method (\backtrans{}) outperforms the generation of pseudo data from raw grammatical sentences (\directnoise{}).
By contrast, the current state of the art model~\citep{zhao2019improving} uses the \directnoise{} method.

In this study, we investigate these decisions regarding pseudo data, our goal being to provide the research community with an improved understanding of the incorporation of pseudo data.
Through extensive experiments, we determine suitable settings for GEC. %
We justify the reliability of the proposed settings by demonstrating their strong performance on benchmark datasets.
Specifically, without any task-specific techniques or architecture, our model outperforms not only all previous single-model results but also all ensemble results except for the ensemble result by~\citet{grundkiewicz:2019:bea}\footnote{The paper (Grundkiewicz et al. 2019) has not been published yet at the time of submission.}.
By applying task-specific techniques, we further improve the performance and achieve state-of-the-art performance on the CoNLL-2014 test set and the official test set of the BEA-2019 shared task.

\section{Problem Formulation and Notation}
\label{sec:problem-formulation}

In this section, we formally define the GEC task discussed in this paper.
Let $\mathcal{D}$ be the GEC training data that comprise pairs of an ungrammatical source sentence $\bm{X}$ and grammatical target sentence $\bm{Y}$, i.e., $\mathcal{D}=\{(\bm{X}_n, \bm{Y}_n)\}_{n}$. 
Here, $\lvert\mathcal{D}\rvert$ denotes the number of sentence pairs in the dataset $\mathcal{D}$.

Let $\bm{\Theta}$ represent all trainable parameters of the model.
Our objective is to find the optimal parameter set $\widehat{\bm{\Theta}}$ that minimizes the following objective function $\mathcal{L}(\mathcal{D}, \bm{\Theta})$ for the given training data $\mathcal{D}$:
\begin{align}
\mathcal{L}(\mathcal{D}, \bm{\Theta}) &= - \frac{1}{\vert \mathcal{D} \vert} \sum_{(\bm{X}, \bm{Y}) \in \mathcal{D}}\!\!\!\log(p( \bm{Y} | \bm{X}, \bm{\Theta})),
\label{eq:loss1}
\end{align}
where $p( \bm{Y} | \bm{X}, \bm{\Theta})$ denotes the conditional probability of $\bm{Y}$ given $\bm{X}$.

In the standard supervised learning setting, the parallel data $\mathcal{D}$ comprise only ``genuine'' parallel data \genuinedata{} (i.e., $\mathcal{D}=\genuinedata{}$).
However, in GEC, incorporating pseudo data \pseudodata{} that are generated from grammatical sentences $\bm{Y} \in \mathcal{T}$, where \seedcorpus{} represents \textit{seed corpus} (i.e., a set of grammatical sentences), is common~\citep{xie:2018:NAACL,zhao2019improving,grundkiewicz:2019:bea}.

Our interest lies in the following three nontrivial aspects of Equation~\ref{eq:loss1}.
\textbf{Aspect (i)}: multiple methods for generating pseudo data \pseudodata{} are available (Section~\ref{sec:noising_methods}).
\textbf{Aspect (ii)}: options for the seed corpus \seedcorpus{} are numerous.
To the best of our knowledge, how the seed corpus domain affects the model performance is yet to be shown.
We compare three corpora, namely, Wikipedia, Simple Wikipedia (SimpleWiki) and English Gigaword, as a first trial.
Wikipedia and SimpleWiki have similar domains, but different grammatical complexities.
Therefore, we can investigate how grammatical complexity affects model performance by comparing these two corpora.
We assume that Gigaword contains the smallest amount of noise among the three corpora.
We can therefore use Gigaword to investigate whether clean text improves model performance.
\textbf{Aspect (iii)}: at least two major settings for incorporating \pseudodata{} into the optimization of Equation~\ref{eq:loss1} are available.
One is to use the two datasets jointly by concatenating them as $\mathcal{D} = \genuinedata{} \cup \pseudodata{}$, which hereinafter we refer to as \joint{}.
The other is to use \pseudodata{} for pretraining, namely, minimizing $\mathcal{L}(\pseudodata{}, \bm{\Theta})$ to acquire $\bm{\Theta}^{\prime}$, and then fine-tuning the model by minimizing $\mathcal{L}(\genuinedata{}, \bm{\Theta}^{\prime})$; hereinafter, we refer to this setting as \pretrain{}.
We investigate these aspects through our extensive experiments (Section~\ref{sec:experiment}).

\section{Methods for Generating Pseudo Data}
\label{sec:noising_methods}
In this section, we describe three methods for generating pseudo data.
In Section~\ref{sec:experiment}, we experimentally compare these methods. %

\noindent\textbf{\backtrans{} and \samplebacktrans{}}\hspace*{3mm} 
Backtranslation for the EncDec model was proposed originally by~\citet{sennrich:2016:backtrans}.
In backtranslation, a reverse model, which generates an ungrammatical sentence from a given grammatical sentence, is trained.
The output of the reverse model is paired with the input and then used as pseudo data.

\backtrans{} is a variant of backtranslation that was proposed by \citet{xie:2018:NAACL}\footnote{referred as ``random noising'' in \citet{xie:2018:NAACL}}.
This method adds $r\betanoise{}$ to the score of each hypothesis in the beam for every time step.
Here, noise $r$ is sampled uniformly from the interval $[0, 1]$, and $\betanoise{} \in \mathbb{R}_{\geq 0}$ is a hyper-parameter that controls the noise scale.
If we set $\betanoise{}=0$, then \backtrans{} is identical to standard backtranslation.

\samplebacktrans{} is another variant of backtranslation, which was proposed by \citet{edunov:2018:EMNLP} for MT.
In \samplebacktrans{}, sentences are decoded by sampling from the distribution of the reverse model.

\noindent\textbf{\directnoise{}}\hspace*{3mm} 
Whereas \backtrans{} and \samplebacktrans{} generate ungrammatical sentences with a reverse model, \directnoise{} injects noise ``directly'' into grammatical sentences~\citep{edunov:2018:EMNLP,zhao2019improving}.
Specifically, for each token in the given sentence, this method probabilistically chooses one of the following operations: (i) masking with a placeholder token \mask{}, (ii) deletion, (iii) insertion of a random token, and (iv) keeping the original\footnote{The detailed algorithm is described in Appendix~\ref{appendix:algorithm}.}.
For each token, the choice is made based on the categorical distribution ($\mu_{\mathrm{mask}}, \mu_{\mathrm{deletion}}, \mu_{\mathrm{insertion}}, \mu_{\mathrm{keep}}$).

\section{Experiments}
\label{sec:experiment}
The goal of our experiments is to investigate \textbf{aspect (i)--(iii)} introduced in Section~\ref{sec:problem-formulation}.
To ensure that the experimental findings are applicable to GEC in general, we design our experiments by using the following two strategies:
(i) we use an off-the-shelf EncDec model without any task-specific architecture or techniques; (ii) we conduct hyper-parameter tuning, evaluation and comparison of each method or setting on the validation set.
At the end of experiments (Section~\ref{subsec:comparison-with-other-studies}), we summarize our findings and propose suitable settings.
We then perform a single-shot evaluation of their performance on the test set.

\subsection{Experimental Configurations}
\label{subsec:experimental-configurations}
\noindent\textbf{Dataset}\hspace*{3mm}
The BEA-2019 workshop official dataset\footnote{Details of the dataset is in Appendix~\ref{appendix:bea-2019-dataset}.} is the origin of the training and validation data of our experiments.
Hereinafter, we refer to the training data as BEA-train.
We create validation data (BEA-valid) by randomly sampling sentence pairs from the official validation split\footnote{The detailed data preparation process is in Appendix~\ref{appendix:data-preparation-process}.}.

As a seed corpus \seedcorpus{}, we use SimpleWiki\footnote{\url{https://simple.wikipedia.org}}, Wikipedia\footnote{We used \texttt{2019-02-25} dump file at \url{https://dumps.wikimedia.org/other/cirrussearch/}.} or Gigaword\footnote{We used the English Gigaword Fifth Edition (LDC Catalog No.: LDC2011T07).}.
We apply the noizing methods described in Section~\ref{sec:noising_methods} to each corpus and generate pseudo data \pseudodata{}.
The characteristics of each dataset are summarized in Table~\ref{table:data_summary}.

\begin{table}[t]
\centering
\scriptsize
\tabcolsep 1mm
\begin{tabular}{lrccc}
\toprule
Dataset          & \#sent (pairs) & \#refs. & Split & Scorer \\
\midrule
BEA-train        & 561,410 &  1      & train  & - \\
BEA-valid        & 2,377   &  1      & valid  & ERRANT \\
\midrule
CoNLL-2014       & 1,312   &  2      & test   & ERRANT \& $M^2$~scorer\\
JFLEG            & 1,951   &  4      & test   & GLEU \\
BEA-test         & 4,477   &  5      & test   & ERRANT \\
\midrule
SimpleWiki${}^{*}$    & 1,369,460   &  -    & -  & - \\
Wikipedia${}^{*}$     & 145,883,941 &  -    & -  & - \\
Gigaword${}^{*}$      & 131,864,979 &  -    & -  & - \\
\bottomrule
\end{tabular}
\vskip -2mm
\caption{Summary of datasets used in our experiments. Dataset marked with ``*'' is a seed corpus \seedcorpus{}.}
\label{table:data_summary}
\end{table}

\noindent\textbf{Evaluation}\hspace*{3mm}
We report results on BEA-valid, the official test set of the BEA-2019 shared task (BEA-test), the CoNLL-2014 test set (CoNLL-2014)~\citep{ng:2014:conll}, and the JFLEG test set (JFLEG)~\citep{napoles:2017:EACL}.
All reported results (except ensemble) are the average of five distinct trials using five different random seeds.
We report the scores measured by ERRANT~\citep{bryant:2017:automatic,felice:2016:automatic} for BEA-valid, BEA-test, and CoNLL-2014.
As the reference sentences of BEA-test are publicly unavailable, we evaluate the model outputs on \texttt{CodaLab}\footnote{\url{https://competitions.codalab.org/competitions/20228}} for BEA-test.
We also report results measured by the $M^2$~scorer~\citep{dahlmeier:2012:M2} on CoNLL-2014 to compare them with those of previous studies.
We use the GLEU metric~\citep{napoles:2015:ACL,napoles:2016:gleu} for JFLEG.

\noindent\textbf{Model}\hspace*{3mm}
We adopt the \textit{Transformer} EncDec model~\citep{vaswani:2017:NIPS} using the \texttt{fairseq} toolkit~\citep{ott2019:arxiv:fairseq} and use the ``Transformer (big)'' settings of~\citet{vaswani:2017:NIPS}. %

\noindent\textbf{Optimization}\hspace*{3mm}
For the \joint{} setting, we optimize the model with Adam~\cite{kingma:2015:ICLR}.
For the \pretrain{} setting, we pretrain the model with Adam and then fine-tune it on BEA-train using Adafactor~\citep{shazeer:2018:adafactor}\footnote{The detailed hyper-parameters are listed in Appendix~\ref{appendix:hyper-parameter-settings}.}.

\begin{table}[t!]
\centering
\small
\begin{tabular}{lccc}
\toprule
Method  & \multicolumn{1}{c}{Prec.} & \multicolumn{1}{c}{Rec.} & \multicolumn{1}{c}{\fscore{}} \\
\midrule
Baseline                          &  46.6  &  23.1  &  38.8  \\
\midrule
\samplebacktrans{}                &  44.6  &  27.4  &  39.6 \\
\backtrans{}                      &  42.5  &  \textbf{31.3}  &  39.7 \\
\directnoise{}                    &  \textbf{48.9}  &  25.7  &  \textbf{41.4} \\
\bottomrule
\end{tabular}

\vskip -2mm
\caption{Performance of models on BEA-valid: a value in \textbf{bold} indicates the best result within the column. The seed corpus \seedcorpus{} is SimpleWiki.}
\label{table:result-pseudo-data-generation}
\end{table}

\subsection{Aspect (i): Pseudo Data Generation}
\label{subsec:pseudo-data-generation-methods}

We compare the effectiveness of the \backtrans{}, \samplebacktrans{}, and \directnoise{} methods for generating pseudo data.
In \directnoise{}, we set $(\mu_{\mathrm{mask}}, \mu_{\mathrm{deletion}}, \mu_{\mathrm{insertion}}, \mu_{\mathrm{keep}})\!=\!(0.5, 0.15, 0.15, 0.2)$\footnote{These values are derived from preliminary experiments (Appendix~\ref{appendix:mask-probability-of-directnoise}).}.
We use $\betanoise{}\!=\!6$ for \backtrans{}\footnote{$\betanoise{}=6$ achieved the best \fscore{} in our preliminary experiments (Appendix~\ref{appendix:noise-strength-of-backtrans}).}.
In addition, we use (i) the \joint{}  setting and (ii) all of SimpleWiki as the seed corpus \seedcorpus{} throughout this section.

The results are summarized in Table~\ref{table:result-pseudo-data-generation}.
\backtrans{} and \samplebacktrans{} show competitive values of \fscore{}.
Given this result, we exclusively use \backtrans{} and discard \samplebacktrans{} for the rest of the experiments.
The advantage of \backtrans{} is that its effectiveness in GEC has already been demonstrated by~\citet{xie:2018:NAACL}.
In addition, in our preliminary experiment, \backtrans{} decoded ungrammatical sentence 1.2 times faster than \samplebacktrans{} did.
We also use \directnoise{} because it achieved the best value of \fscore{} among all the methods.

\subsection{Aspect (ii): Seed Corpus \seedcorpus{}}
\label{subsec:source-of-pseudo-data}
We investigate the effectiveness of the seed corpus \seedcorpus{} for generating pseudo data \pseudodata{}.
The three corpora (Wikipedia, SimpleWiki and Gigaword) are compared in Table~\ref{table:result-data-source-for-pseudo}.
We set $\lvert\pseudodata{}\rvert=1.4\mathrm{M}$.
The difference in \fscore{} is small, which implies that the seed corpus \seedcorpus{} has only a minor effect on the model performance.
Nevertheless, Gigaword consistently outperforms the other two corpora.
In particular, \directnoise{} with Gigaword achieves the best value of \fscore{} among all the configurations.

\begin{table}[t!]
\centering
\small
\tabcolsep 1.5mm
\begin{tabular}{@{}lcccc@{}}
\toprule
Method           & Seed Corpus \seedcorpus{}  & Prec.  & Rec.  & \fscore{} \\
\midrule
Baseline         & N/A            &  46.6  &  23.1  &  38.8  \\
\midrule
\backtrans{}     & Wikipedia      &  43.8  &  30.8  &  40.4  \\
\backtrans{}     & SimpleWiki     &  42.5  &  31.3  &  39.7  \\
\backtrans{}     & Gigaword       &  43.1  &  33.1  &  40.6  \\
\midrule
\directnoise{}   & Wikipedia      &  48.3  &  25.5  &  41.0  \\
\directnoise{}   & SimpleWiki     &  48.9  &  25.7  &  41.4  \\
\directnoise{}   & Gigaword       &  48.3  &  26.9  &  41.7 \\
\bottomrule
\end{tabular}
\vskip -2mm
\caption{Performance on BEA-valid when changing the seed corpus \seedcorpus{} used for generating pseudo data ($\lvert\pseudodata{}\rvert=1.4\mathrm{M}$).}
\label{table:result-data-source-for-pseudo}
\end{table}

\subsection{Aspect (iii): Optimization Setting}
\label{subsec:optimization-strategy}
We compare the \joint{} and \pretrain{} optimization settings.
We are interested in how each setting performs when the scale of the pseudo data \pseudodata{} compared with that of the genuine parallel data \genuinedata{} is (i) approximately the same ($\lvert\mathcal{D}_{p}\rvert=1.4\mathrm{M}$) and (ii) substantially bigger ($\lvert\mathcal{D}_{p}\rvert=14\mathrm{M}$).
Here, we use Wikipedia as the seed corpus \seedcorpus{} instead of SimpleWiki or Gigaword for two reasons. 
First, SimpleWiki is too small for the experiment \textbf{(b)} (see Table~\ref{table:data_summary}).
Second, the fact that Gigaword is not freely available makes it difficult for other researchers to replicate our results.

\noindent\textbf{(a) Joint Training or Pretraining}\hspace*{3mm}
Table~\ref{table:joint-or-pretrain} presents the results.
The most notable result here is that \pretrain{} demonstrates the properties of \textit{more pseudo data and better performance}, whereas \joint{} does not.
For example, in \backtrans{}, increasing $\lvert \pseudodata{} \rvert$ (1.4M $\rightarrow$ 14M) improves \fscore{} on \pretrain{} ($41.1 \rightarrow 44.5$).
By contrast, \fscore{} does not improve on \joint{} ($40.4 \rightarrow 40.3$).
An intuitive explanation for this case is that when pseudo data \pseudodata{} are substantially more than genuine data \genuinedata{}, the teaching signal from \pseudodata{} becomes dominant in \joint{}.
\pretrain{} alleviates this problem because the model is trained with only \genuinedata{} during fine-tuning.
We therefore suppose that \pretrain{} is crucial for utilizing extensive pseudo data.

\begin{table}[t!]
\centering
\small
\tabcolsep 1.5pt
\begin{tabular}{llcccc}
\toprule
Optimization & Method & $\lvert\pseudodata{}\rvert$ & \multicolumn{1}{c}{Prec.} & \multicolumn{1}{c}{Rec.} & \multicolumn{1}{c}{\fscore{}}  \\
\midrule
N/A         & Baseline        & 0    &  46.6  &  23.1  &  38.8  \\
\midrule
\pretrain{} &\backtrans{}     & 1.4M &  49.6  &  24.3  &  41.1  \\
\pretrain{} &\directnoise{}   & 1.4M &  48.4  &  21.2  &  38.5  \\
\joint{}    &\backtrans{}     & 1.4M &  43.8  &  30.8  &  40.4  \\
\joint{}    &\directnoise{}   & 1.4M &  48.3  &  25.5  &  41.0  \\
\midrule
\pretrain{} &\backtrans{}     & 14M  &  50.6  &  30.1  &  44.5  \\
\pretrain{} &\directnoise{}   & 14M  &  49.8  &  25.8  &  42.0  \\
\joint{}    &\backtrans{}     & 14M  &  43.0  &  32.3  &  40.3  \\
\joint{}    &\directnoise{}   & 14M  &  48.7  &  23.5  &  40.1  \\
\bottomrule
\end{tabular}
\vskip -2mm
\caption{Performance of the model with different optimization settings on BEA-valid. The seed corpus \seedcorpus{} is Wikipedia.}
\label{table:joint-or-pretrain}
\end{table}

\begin{figure}[t!]
  \center
  \includegraphics[width=\hsize]{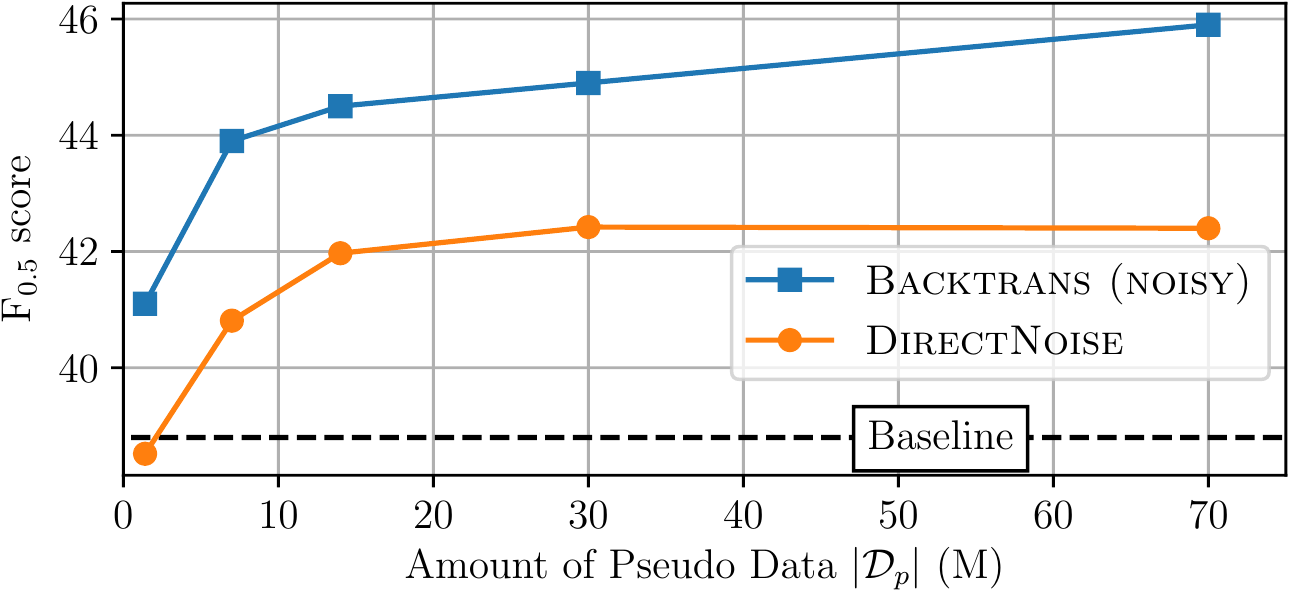}
  \vskip -3mm
  \caption{Performance on BEA-valid for different amounts of pseudo data ($\lvert\pseudodata{}\rvert$). The seed corpus \seedcorpus{} is Wikipedia.}
  \label{fig:amount-of-pseudo-data}
\end{figure}

\noindent\textbf{(b) Amount of Pseudo Data}\hspace*{3mm}
We investigate how increasing the amount of pseudo data affects the \pretrain{} setting.
We pretrain the model with different amounts of pseudo data \{1.4M, 7M, 14M, 30M, 70M\}.
The results in Figure~\ref{fig:amount-of-pseudo-data} show that \backtrans{} has superior sample efficiency to \directnoise{}.
The best model (pretrained with 70M \backtrans{}) achieves $\fscore{}\!=\!45.9$.

\subsection{Comparison with Current Top Models}
\label{subsec:comparison-with-other-studies}

\begin{table*}[t!]
\centering
\scriptsize
\begin{tabular}{lccccccccccc}
\toprule
 & & \multicolumn{3}{c}{\begin{tabular}{c} CoNLL-2014 \\ ($M^2$ scorer) \end{tabular}} & \multicolumn{3}{c}{\begin{tabular}{c} CoNLL-2014 \\ (ERRANT) \end{tabular}}& JFLEG & \multicolumn{3}{c}{\begin{tabular}{c} BEA-test \\ (ERRANT) \end{tabular}} \\
\cmidrule(r){3-5}\cmidrule(r){6-8}\cmidrule(r){9-9}\cmidrule(){10-12}
\multicolumn{1}{c}{Model}  & Ensemble   & Prec.     & Rec.     & \fscore{} & Prec.     & Rec.     & \fscore{}   & GLEU & Prec. & Rec. & \fscore{} \\
\midrule
 \citet{chollampatt:2018:AAAI}   &     & 60.9      & 23.7     & 46.4 & -      & -     & -   & 51.3 & -      & -     & - \\
 \citet{junczys:2018:NAACL}      &     & -         & -        & 53.0 & -      & -     & -   & 57.9 & -      & -     & - \\
 \citet{grundkiewicz:2018:NAACL} &          & 66.8 & 34.5   & 56.3 & -      & -     & -   & 61.5 & -      & -     & - \\
 \citet{lichtarge2019corpora}    &      & 65.5     & 37.1     & 56.8 & -      & -     & -   & 61.6 & -      & -     & - \\
\midrule
 \citet{chollampatt:2018:AAAI} &  \checkmark   & 65.5      & 33.1     & 54.8 & -      & -     & -   & 57.5& -      & -     & -  \\
 \citet{junczys:2018:NAACL}    &  \checkmark   & 61.9      & 40.2     & 55.8 & -      & -     & -   & 59.9& -      & -     & -  \\
 \citet{lichtarge2019corpora}  &  \checkmark   & 66.7      & 43.9     & 60.4 & -     & -     & -   & \textbf{63.3}& -      & -     & -  \\
 \citet{zhao2019improving}     &  \checkmark   & 71.6  & 38.7     & 61.2 & -     & -     & -   & 61.0 & -      & -     & - \\
 \citet{grundkiewicz:2019:bea} &  \checkmark   &   -   &  -  & 64.2  & -     & -     & -   & 61.2 & 72.3 & 60.1 & 69.5 \\
\midrule
\midrule
\textsc{PretLarge}             &   &  67.9  &  44.1  &  61.3 &  61.2  &  42.0  &  56.0   &  59.7 & 65.5 & 59.4 & 64.2  \\  
\midrule
\textsc{PretLarge}+\spell{}+\textsc{R2L}  & \checkmark &  72.4  &  \textbf{46.1}  &  \textbf{65.0}  & 67.3&  \textbf{44.0} & \textbf{60.9} &   61.4  & 72.1 & \textbf{61.8} & 69.8 \\

\textsc{PretLarge}+\spell{}+\textsc{R2L}+\textsc{SED}  &  \checkmark   &   \textbf{73.3}   &  44.2   & 64.7  &    \textbf{68.1} & 42.1 &   60.6  &  61.2 & \textbf{74.7} & 56.7 & \textbf{70.2} \\
\bottomrule
\end{tabular}
\vskip -2mm
\caption{Comparison of our best model and current top models: a \textbf{bold} value indicates the best result within the column.}
\label{tab:sota}
\end{table*}

The present experimental results show that the following configurations are effective for improving the model performance: (i) the combination of \joint{} and Gigaword (Section~\ref{subsec:source-of-pseudo-data}), (ii) the amount of pseudo data \pseudodata{} not being too large in \joint{} (Section~\ref{subsec:optimization-strategy}(a)), and (iii) \pretrain{} with \backtrans{} using large pseudo data \pseudodata{} (Section~\ref{subsec:optimization-strategy}(b)).
We summarize these findings and attempt to combine \pretrain{} and \joint{}.
Specifically, we pretrain the model using 70M pseudo data of \backtrans{}.
We then fine-tune the model by combining BEA-train and relatively small \directnoise{} pseudo data generated from Gigaword (we set $\lvert\pseudodata{}\rvert=250\mathrm{K}$).
However, the performance does not improve on BEA-valid.
Therefore, the best approach available is simply to pretrain the model with large (70M) \backtrans{} pseudo data and then fine-tune using BEA-train, which hereinafter we refer to as \textsc{PretLarge}.
We use Gigaword for the seed corpus \seedcorpus{} because it has the best performance in Table~\ref{table:result-data-source-for-pseudo}.

We evaluate the performance of \textsc{PretLarge} on test sets and compare the scores with the current top models.
Table~\ref{tab:sota} shows a remarkable result, that is, our \textsc{PretLarge} achieves $\fscore{}=61.3$ on CoNLL-2014. 
This result outperforms not only all previous single-model results but also all ensemble results except for that by~\citet{grundkiewicz:2019:bea}. %

To further improve the performance, we incorporate the following techniques that are widely used in shared tasks such as BEA-2019 and WMT\footnote{\url{http://www.statmt.org/wmt19/}}:

\noindent\textbf{Synthetic Spelling Error} (\textsc{SSE})\hspace*{3mm} 
\citet{lichtarge2019corpora} proposed the method of probabilistically injecting character-level noise into the source sentence of pseudo data \pseudodata{}.
Specifically, one of the following operations is applied randomly at a rate of 0.003 per character: deletion, insertion, replacement, or transposition of adjacent characters.

\noindent\textbf{Right-to-left Re-ranking} (\textsc{R2L})\hspace*{3mm} 
Following~\citet{sennrich:2016:wmt,sennrich:2017:wmt,grundkiewicz:2019:bea}, we train four right-to-left models.
The ensemble of four left-to-right models generate $n$-best candidates and their corresponding scores (i.e., conditional probabilities).
We then pass each candidate to the ensemble of the four right-to-left models and compute the score.
Finally, we re-rank the $n$-best candidates based on the sum of the two scores.

\noindent\textbf{Sentence-level Error Detection} (\textsc{SED})\hspace*{3mm}
\textsc{SED} classifies whether a given sentence contains a grammatical error.
\citet{asano:2019:bea} proposed incorporating \textsc{SED} into the evaluation pipeline and reported improved precision.
Here, the GEC model is applied only if SED detects a grammatical error in the given source sentence.
The motivation is that \textsc{SED} could potentially reduce the number of false-positive errors of the GEC model.
We use the re-implementation of the BERT-based SED model~\citep{asano:2019:bea}.

Table~\ref{tab:sota} presents the results of applying \textsc{SSE}, \textsc{R2L}, and \textsc{SED}.
It is noteworthy that \textsc{PretLarge}+\spell{}+\textsc{R2L} achieves state-of-the-art performance on both CoNLL-2014 ($\fscore{}=65.0$) and BEA-test ($\fscore{}=69.8$), which are better than those of the best system of the BEA-2019 shared task~\citep{grundkiewicz:2019:bea}.
In addition, \textsc{PretLarge}+\spell{}+\textsc{R2L}+\textsc{SED} can further improve the performance on BEA-test ($\fscore{}=70.2$).
However, unfortunately, incorporating \textsc{SED} decreased the performance on CoNLL-2014 and JFLEG.
This fact implies that \textsc{SED} is sensitive to the domain of the test set since the \textsc{SED} model is fine-tuned with the official validation split of BEA dataset.
We leave this sensitivity issue as our future work.

\section{Conclusion}
In this study, we investigated several aspects of incorporating pseudo data for GEC.
Through extensive experiments, we found the following to be effective:
(i) utilizing Gigaword as the seed corpus, and
(ii) pretraining the model with \backtrans{} data.
Based on these findings, we proposed suitable settings for GEC.
We demonstrated the effectiveness of our proposal by achieving state-of-the-art performance on the CoNLL-2014 test set and the BEA-2019 test set.

\section*{Acknowledgements}
We thank the three anonymous reviewers for their insightful comments.
We are deeply grateful to Takumi Ito and Tatsuki Kuribayashi for kindly sharing the re-implementation of \backtrans{}.
The work of Jun Suzuki was supported in part by JSPS KAKENHI Grant Number JP19104418 and AIRPF Grant Number 30AI036-8.

\bibliography{references}
\bibliographystyle{acl_natbib}

\newpage
\onecolumn
\appendix
\section{\directnoise{} Algorithm}
\label{appendix:algorithm}

The \directnoise{} algorithm is described in Algorithm~\ref{alg:corpus_generation}
Here, $\bm{X}$ consists of sequence of $I$ tokens, namely, $\bm{X}=(x_1,\dots,x_{I})$ where $x_i$ denotes $i$-th token of $\bm{X}$.
Similarly, $\bm{Y}$ consists of sequence of $J$ tokens, namely, $\bm{Y}=(y_1,\dots,y_{J})$ where $y_j$ denotes $j$-th token of $\bm{Y}$.

\begin{algorithm}[h!]
  \small
  \DontPrintSemicolon
  \KwData{Grammatical sentence $\bm{Y} \in \mathcal{T}$}
  \KwResult{Pseudo Corpus \pseudodata{}}
   $\mathcal{D}_{p} = \{\}$\tcp*{create empty set}
   $\bm{\mu} = \{\mu_{\mathrm{mask}}, \mu_{\mathrm{deletion}}, \mu_{\mathrm{insertion}}, \mu_{\mathrm{keep}} \}~\mbox{s.t.}~\Sigma\bm{\mu}\!=\!1$ \;
  \For{$\bm{Y} \in \mathcal{T}$}{
     $\bm{X} = (~)$ \;
    \For{$j \in (1,\dots,J)$} {
     $action \sim Cat(action | \bm{\mu})$ \;
    \If{action is keep}{
      append $y_j$ to $\bm{X}$
    }
    \ElseIf{action is mask}{
       append $\mbox{\mask{}})$ to $\bm{X}$
    }
     \ElseIf{action is deletion} {
      continue
    }
     \ElseIf{action is insertion} {
     append $y_j$ to $\bm{X}$\;
     $w = sample\_from\_unigram\_distribution(\genuinedata{})$ \;
     append $w$ to $\bm{X}$\;
    }
    }
    $\pseudodata{} = \pseudodata{} \cup \{(\bm{X}, \bm{Y})\}$
  }
	\caption{\directnoise{} Algorithm}
  \label{alg:corpus_generation}
\end{algorithm}

\newpage
\section{BEA-2019 Workshop Official Dataset}
\label{appendix:bea-2019-dataset}
The BEA-2019 Workshop official dataset consists of following corpora: the First Certificate in English corpus~\citep{Yannakoudakis:11:ACL}, Lang-8 Corpus of Learner English (Lang-8)~\citep{mizumoto:2011:IJCNLP,Tajiri:12:ACL}, the National University of Singapore Corpus of Learner English (NUCLE)~\citep{Dahlmeier:13:BEA}, and W\&I+LOCNESS~\citep{Yannakoudakis:18:Journal,granger:1998:LEC}. 
The data is publicly available at \url{https://www.cl.cam.ac.uk/research/nl/bea2019st/}.

\section{Data Preparation Process}
\label{appendix:data-preparation-process}
The training data (BEA-train) is tokenized using \texttt{spaCy} tokenizer\footnote{\url{https://spacy.io/}}.
We used \texttt{en\_core\_web\_sm-2.1.0} model\footnote{\url{https://github.com/explosion/spacy-models/releases/tag/en_core_web_sm-2.1.0}}.
We remove sentence pairs that have identical source and target sentences from the training set, following~\citep{chollampatt:2018:AAAI}.
Then we acquire subwords from target sentence through byte-pair-encoding (BPE)~\citep{sennrich:2016:ACL} algorithm.
We used \texttt{subword-nmt} implementation\footnote{\url{https://github.com/rsennrich/subword-nmt}}.
We apply BPE splitting to both source and target text.
The number of merge operation is set to 8,000.

\newpage
\section{Hyper-parameter Settings}
\label{appendix:hyper-parameter-settings}

\begin{table}[h]
\centering
\small
\begin{tabular}{@{}lp{105mm}@{}}
\toprule
   Configurations         &   Values \\ \midrule
   Model Architecture     &   Transformer~\citep{vaswani:2017:NIPS} (``big'' setting)    \\
   Optimizer              &   Adam~\citep{kingma:2015:ICLR}  ($\beta_{1}=0.9, \beta_{2}=0.98, \epsilon=1\times10^{-8}$)  \\
   Learning Rate Schedule &   Same as described in Section 5.3 of \citet{vaswani:2017:NIPS}     \\
   Number of Epochs       &   40   \\
   Dropout                &   0.3  \\
   Stopping Criterion     &   Train model for 40 epochs. During the training, save model parameter for every 500 updates. Then take average of last 20 checkpoints.     \\ 
   Gradient Clipping      &   1.0  \\
   Loss Function          &   Label smoothed cross entropy (smoothing value: $\epsilon_{ls}=0.1$)~\citep{szegedy:2016:rethinking}     \\
   Beam Search            &   Beam size 5 with length-normalization    \\
   \bottomrule
\end{tabular}
\caption{Hyper-parameter for \joint{} optimization}
\label{tab:my-table}
\end{table}

\begin{table}[h]
\centering
\small
\begin{tabular}{@{}ll@{}}
\toprule
Configurations & Values \\ \midrule
\multicolumn{2}{c}{\textbf{Pretraining}} \\
\midrule
   Model Architecture     &    Transformer~\citep{vaswani:2017:NIPS} (``big'' setting)    \\
   Optimizer              &    Adam~\citep{kingma:2015:ICLR}  ($\beta_{1}=0.9, \beta_{2}=0.98, \epsilon=1\times10^{-8}$)  \\
   Learning Rate Schedule &    Same as described in Section 5.3 of \citet{vaswani:2017:NIPS}     \\
   Number of Epochs       &    10   \\
   Dropout                &    0.3    \\
   Gradient Clipping      &    1.0    \\
   Loss Function          &    Label smoothed cross entropy (smoothing value: $\epsilon_{ls}=0.1$)~\citep{szegedy:2016:rethinking}     \\
\midrule
\multicolumn{2}{c}{\textbf{Fine-tuning}} \\
\midrule
   Model Architecture     &   Transformer~\citep{vaswani:2017:NIPS} (``big'' setting)    \\
   Optimizer              &   Adafactor~\citep{shazeer:2018:adafactor} \\
   Learning Rate Schedule &   Constant learning rate of $3\times10^{-5}$  \\
   Number of Epochs       &   30   \\
   Dropout                &   0.3    \\
   Stopping Criterion     &   Use the model with the best validation perplexity on BEA-valid \\
   Gradient Clipping      &   1.0    \\
   Loss Function          &   Label smoothed cross entropy (smoothing value: $\epsilon_{ls}=0.1$)~\citep{szegedy:2016:rethinking}     \\
   Beam Search            &   Beam size 5 with length-normalization    \\
  \bottomrule
\end{tabular}
\caption{Hyper-parameter for \pretrain{} optimization}
\label{tab:my-table2}
\end{table}

\newpage
\section{Mask Probability of \directnoise{}}
\label{appendix:mask-probability-of-directnoise}
In this paper, we exclusively focused on the effectiveness of $\mu_{mask}$, and therefore we deliberately fixed $\mu_{\mathrm{keep}}=0.2$, and used $\mu_{\mathrm{insertion}}=\mu_{\mathrm{deletion}}=(1-\mu_{\mathrm{keep}}-\mu_{\mathrm{mask}}) /2$

\begin{figure}[h]
  \center
  \includegraphics[width=\hsize]{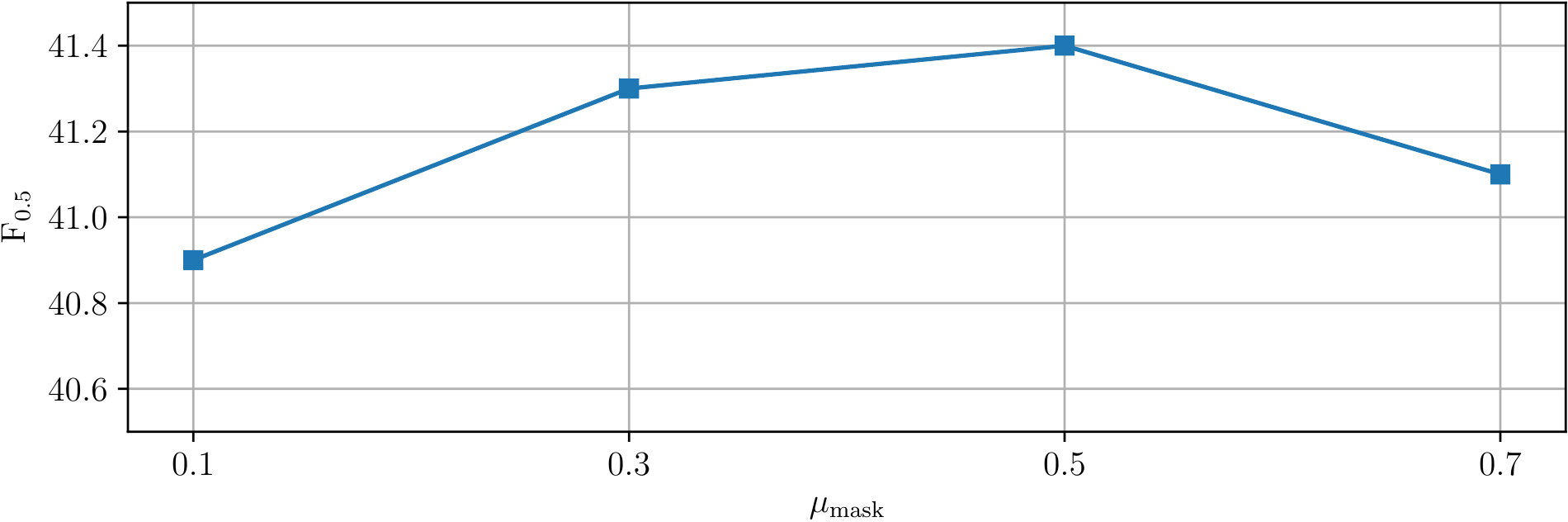}
  \caption{Performance of the model on BEA-valid as parameter of \directnoise{} ($\mu_{\mathrm{mask}}$) is varied.}
  \label{fig:valid-f-score-mask}
\end{figure}

We investigated the effectiveness of changing mask probability $\mu_{\mathrm{mask}}$ of \backtrans{} by evaluating the model performance on BEA-valid.
We used entire SimpleWiki as the seed corpus \seedcorpus{}.
The result is summarized in Figure~\ref{fig:valid-f-score-mask}.
Here, increasing $\mu_{\mathrm{mask}}$ within the range of $0.1 < \mu_{\mathrm{mask}} < 0.5$ slightly improved the performance.
Thus, used $\mu_{\mathrm{mask}}=0.5$ in the experiment (Section~\ref{sec:experiment}).

\section{Noise Strength of \backtrans{}}
\label{appendix:noise-strength-of-backtrans}
\begin{figure}[h]
  \center
  \includegraphics[width=\hsize]{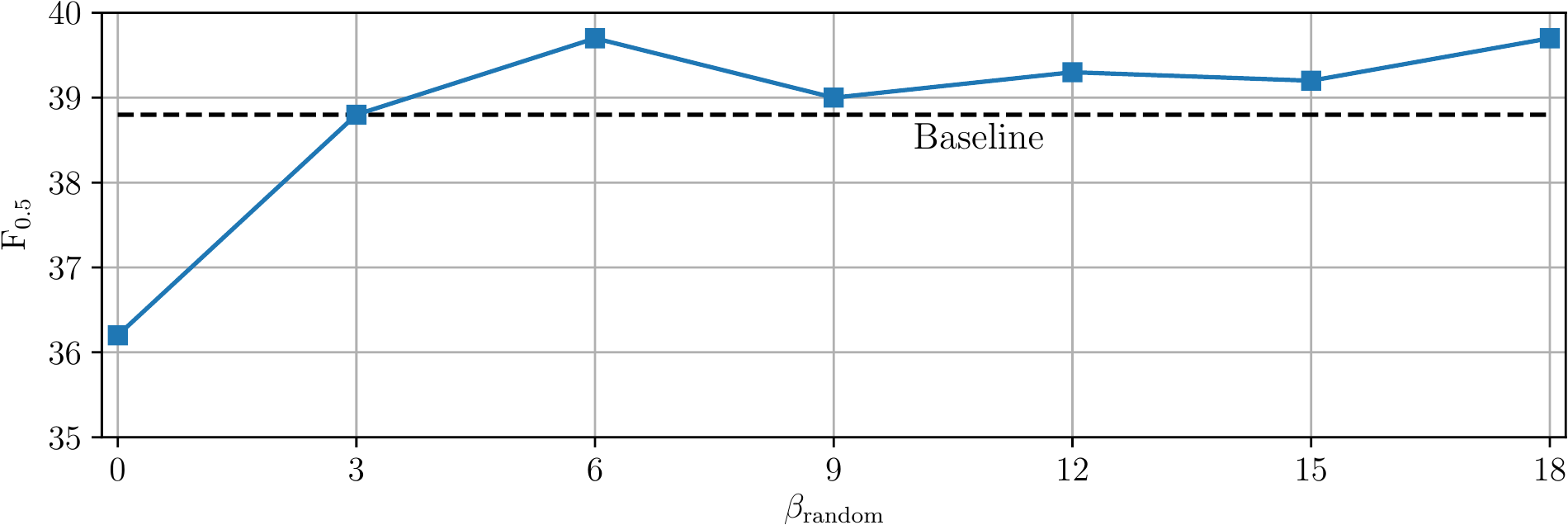}
  \caption{Performance of the model on BEA-valid as parameter of \backtrans{} (\betanoise{}) is varied.}
  \label{fig:valid-f-score-backtrans}
\end{figure}

We investigated the effectiveness of varying \betanoise{} hyper-parameter of \backtrans{} by evaluating its performance on BEA-valid (Figure~\ref{fig:valid-f-score-backtrans}).
We used entire SimpleWiki as the seed corpus \seedcorpus{}.
The figure shows that the performance of backtranslation without noise ($\betanoise{}=0$) is worse than the baseline.
We believe that when there is no noise, reverse-model becomes too conservative to generate grammatical error, as discussed by~\citet{xie:2018:NAACL}.
Thus, the generated pseudo data cannot provide useful teaching signal for the model.

In terms of the scale of the noise, $\betanoise{}=6$ is the best value for \backtrans{}.
Thus, we used this value in the experiment (Section~\ref{sec:experiment}).

\newpage
\section{Examples of Noisy Sentences}
Figure~\ref{fig:noise-example-appendix} shows examples of noisy sentences that are generated by \backtrans{} and \directnoise{}.

\begin{figure*}[h]
\centering
\tabcolsep=1pt
\small
\begin{tabular}{lp{110mm}}
  \toprule
   Original: & \texttt{He died there , but the death date is not clear .} \\
   \backtrans{}: & \texttt{He died at there , but death date is not clear .} \\
   \directnoise{}: & \texttt{\mask{} \mask{} \mask{} , 2 but \mask{} \mask{} \mask{} is not \mask{} \mask{}} \\
   \midrule
   Original: & \texttt{On seeing her his joy knew no bounds .}\\
   \backtrans{}: & \texttt{On seeing her joyful knew no bounds .}\\
   \directnoise{}: & \texttt{\mask{} \mask{} her crahis \mask{} \mask{} \mask{} bke \mask{} .}\\
   \midrule
   Original: & \texttt{Gre@@ en@@ space Information for G@@ rea@@ ter London .} \\
   \backtrans{}: & \texttt{The information for Gre@@ en@@ space information about G@@ rea@@ ter London .} \\
   \directnoise{}: & \texttt{\mask{} \mask{} \mask{} for \mask{} \mask{} \mask{} \mask{}} \\
   \midrule
   Original: & \texttt{The cli@@ p is mixed with images of Toronto streets during power failure .} \\
   \backtrans{}: & \texttt{The cli@@ p is mix with images of Toronto streets during power failure .} \\
   \directnoise{}: & \texttt{The \mask{} is mixed \mask{} images si@@ of The \mask{} streets large \mask{} power R@@ failure place \mask{}} \\
   \midrule
   Original: & \texttt{At the in@@ stitute , she introduced tis@@ sue culture methods that she had learned in the U.@@ S.} \\
   \backtrans{}: & \texttt{At in@@ stitute , She introduced tis@@ sue culture method that she learned in U.@@ S.} \\
   \directnoise{}: & \texttt{\mask{} the the \mask{} \mask{} \mask{} \mask{} tis@@ culture R@@ methods , she P \mask{} the s U.@@ \mask{}} \\
   \bottomrule
   \vspace{-5.0pt}\\
  \end{tabular}
  \vspace{-10.0pt}
\caption{Examples of sentences generated by \backtrans{} and \directnoise{} methods.}
\label{fig:noise-example-appendix}
\end{figure*}

Figure~\ref{fig:direct-noise} shows examples generated by \directnoise{}, when changing the mask probability ($\mu_{\mathrm{mask}}$).

\begin{figure}[h]
\centering
\tabcolsep=1pt
\small
\begin{tabular}{cc}
  \toprule
   $\mu_{\mathrm{mask}}$& Output Sentence\\
  \midrule
   N/A & \texttt{He threw the sand@@ wi@@ ch at his wife .}\\
   \midrule
   0.1 & \texttt{He ale threw , ch his ne@@ wife dar@@ \mask{}}\\
   \midrule
   0.3 & \texttt{\mask{} \mask{} \mask{} \mask{} ch at ament his Research .}\\
   \midrule
   0.5 & \texttt{He o threw the sand@@ ch \mask{} his \mask{} .}\\
   \midrule
   0.7 & \texttt{\mask{} \mask{} sand@@ \mask{} \mask{} \mask{} \mask{} wife \mask{}}\\
   \bottomrule
   \vspace{-5.0pt}\\
  \end{tabular}
  \vspace{-10.0pt}
\caption{Examples generated when varying $\mu_{\mathrm{mask}}$. N/A denotes original text.}
\label{fig:direct-noise}
\end{figure}

\newpage
\section{Performance of the Model without Fine-tuning}
\begin{figure}[h]
  \center
  \includegraphics[width=\hsize]{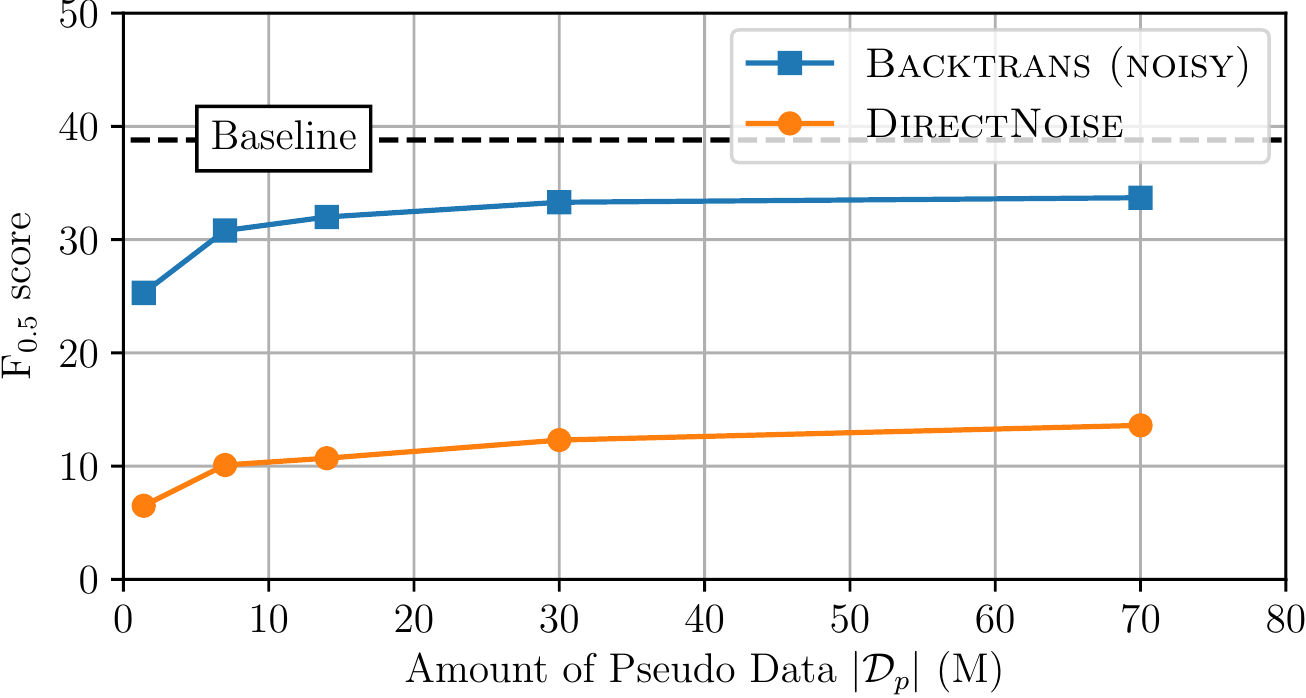}
  \vskip -2mm
  \caption{Performance on BEA-valid when varying the amount of pseudo data ($\lvert\pseudodata{}\rvert$)}
  \label{fig:result-pretrain}
\end{figure}

\pretrain{} setting undergoes two optimization steps, namely, pretraining with pseudo data \pseudodata{} and fine-tuning with genuine parallel data \genuinedata{}.
We report the performance of models with pretraining only (Figure~\ref{fig:result-pretrain}).

\end{document}